\documentclass[11pt,letterpaper]{article}
\usepackage{emnlp2017}
\usepackage{times}
\usepackage{url}
\usepackage{latexsym}
\usepackage{graphicx}

\emnlpfinalcopy 

\title{Native Language Identification on Text and Speech}

\author{Marcos Zampieri\textsuperscript{1}, Alina Maria Ciobanu\textsuperscript{2}, Liviu P. Dinu\textsuperscript{2} \\
  \textsuperscript{1}University of Wolverhampton, United Kingdom\\
  \textsuperscript{2}University of Bucharest, Romania\\
  {\tt marcos.zampieri@uni-koeln.de} \\}

\date{}

\begin{document}
\maketitle
\begin{abstract}
This paper presents an ensemble system combining the output of multiple SVM classifiers to native language identification (NLI). The system was submitted to the NLI Shared Task 2017 fusion track which featured students essays and spoken responses in form of audio transcriptions and iVectors by non-native English speakers of eleven native languages. Our system competed in the challenge under the team name ZCD and was based on an ensemble of SVM classifiers trained on character $n$-grams achieving 83.58\% accuracy and ranking 3\textsuperscript{rd} in the shared task.
\end{abstract}

\section{Introduction}
\label{intro}

Native language identification (NLI) is the task of
automatically identifying non-native speakers' native language based on their foreign language production. As evidenced in \newcite{malmasi2016} NLI is a vibrant research area in NLP and is usually modeled as single-label text classification.

NLI is based on the assumption that the mother tongue influences second language acquisition (SLA) and production. Corpora containing texts and utterances by non-native speakers are used to train systems that are able to recognize features that are prominent in the production of speakers of a particular native language. These features are subsequently used to identify texts (or utterances) that are likely to be written or spoken by speakers of the same language.

There are two important reasons to study NLI. Firstly, there is SLA. NLI methods can be applied to learner corpora to investigate the influence of native language in second language acquisition and production complementing \mbox{corpus-based} and corpus-driven studies. The second reason is a practical one. NLI methods can be an important part of several NLP systems including, for example, author profiling systems developed for forensic linguistics. 

This paper presents the system submitted by the ZCD team to the NLI Shared Task 2017 \cite{nli2017}. The organizers of the challenge provided participants with a dataset containing essays and spoken responses in form of transcriptions and acoustic features (iVectors) by non-native English speakers of eleven native languages taking a standardized assessment of English proficiency for academic purposes. Native languages included are: Arabic, Chinese, French, German, Hindi, Italian, Japanese, Korean, Spanish, Telugu, and Turkish. To discriminate between these eleven native languages we apply an ensemble of multiple linear SVM classifiers trained on character $n$-grams. The main motivation behind the choice of this approach is the success of linear SVMs and SVM ensembles in NLI and in similar text classification tasks such as dialect, language variety, and similar language identification as will be discussed in Section \ref{sec:related}.

\section{Related Work}
\label{sec:related}

\begin{table*}[ht!]
\centering
\scalebox{0.85}{
    \begin{tabular}{lp{10cm}c}
    \hline
    \bf Team  & \bf Approach  & \bf System Paper \\ \hline
    
    Jarvis & SVM trained on character $n$-grams (1-9), word $n$-grams (1-4), and POS $n$-grams (1-4)  & \citep{jarvis-bestgen-pepper:2013:BEA8} \\
    
    Oslo & SVM trained on character $n$-grams (1-7) & \citep{lynum:2013:BEA8} \\
    
    Unibuc & String Kernels and Local Rank Distance (LRD) & \citep{popescu-ionescu:2013:BEA8} \\
    
    MITRE & Bayes ensemble of multiple classifiers  & \citep{henderson-EtAl:2013:BEA8} \\
    
    Tuebingen & SVM trained on word $n$-grams (1-2), and POS $n$-grams (1-5), and syntactic features (dependencies)  & \citep{bykh-EtAl:2013:BEA8} \\
    
    NRC & Ensemble of SVM classifiers trained on character trigrams, word $n$-grams (1-2), POS $n$-grams (2-4), and syntactic features (dependencies) & \citep{goutte-leger-carpuat:2013:BEA8} \\
    
    CMU-Haifa & Maximum Entropy trained on word $n$-grams (1-4), POS $n$-grams (1-4), and spelling features & \citep{tsvetkov-EtAl:2013:BEA8} \\
    
    Cologne-Nijmegen & SVM classifier with TF-IDF weighting trained on character $n$-grams (1-6), word $n$-grams (1-2), and POS $n$-grams (1-4)  & \citep{gebre-EtAl:2013:BEA8} \\
    
    NAIST & SVM trained on character $n$-grams (2-3), word $n$-grams (1-2), and POS $n$-grams (2-3), and syntactic features (dependencies and TSG)  & \citep{mizumoto-EtAl:2013:BEA8} \\
    
    UTD & SVM trained on word $n$-grams (1-2) & \citep{wu-EtAl:2013:BEA8} \\
    \hline
    \end{tabular}
}
\caption{Top ten NLI Shared Task 2013 entries ordered by performance.}
\label{tab:approaches}
\end{table*}

There have been several NLI studies published in the past few years. Due to the availability of suitable language resources for English (e.g. learner corpora), the vast majority of these studies dealt with English \cite{brooke2012measuring,bykh:2014}, however, a few NLI studies have been published on other languages. Examples of NLI applied to languages other than English include Arabic \cite{ionescu2015fast}, Chinese \cite{wang2016your}, and Finnish \cite{malmasi2014finnish}.

To the best of our knowledge, the NLI Shared Task 2013 \cite{nli2013} was the first shared task to provide a benchmark for NLI focusing on written texts by non-native English speakers. A few years later, the 2016 Computational Paralinguistics Challenge \cite{compare2016} included an NLI task on speech data. The NLI Shared Task 2017 combines these two modalities of non-native language production by including essays and spoken responses of test takers in form of transcriptions and iVectors. 

The combination of text and speech has been previously used in similar shared tasks such as the dialect identification shared tasks organized at the VarDial workshop series \cite{vardial2017} and described in more detail in Section \ref{sec:overlap}.

In the next sections we present the most successful entries submitted for the NLI Shared Task 2013 and their overlap with methods applied to dialect, language variety, and similar language identification.

\subsection{NLI Shared Task 2013}

The aforementioned NLI Shared Task 2013 \cite{nli2013} established the first benchmark for NLI on written texts. Organizers of the first NLI task provided participants with the TOEFL 11 \cite{TOEFL11-RR} dataset which contained essays written by students native speakers of the same eleven languages included in the NLI Shared Task 2017. 

Twenty-nine teams participated in the competition, testing a wide range of computational methods for NLI.  In Table \ref{tab:approaches} we list the top ten best entries ranked by performance along with their respective system description papers.

The best system by \newcite{jarvis-bestgen-pepper:2013:BEA8} applied a linear SVM classifier trained on character, word, and POS $n$-grams. Seven out of the ten best entries in the shared task used SVM classifiers. This indicates that SMVs are a very good fit for NLI and motivates us to test SVM classifiers in our ensemble-based system described in this paper.

\subsection{Overlap with Dialect Identification}
\label{sec:overlap}

In the last few years, we observed a significant and important overlap between NLI approaches and computational methods applied to dialect, language variety, and similar language identification. So far the overlap between the two tasks has not been substantially explored in the literature.

Members of several teams that submitted systems to the NLI Shared Task 2013, some of them presented in Table \ref{tab:approaches}, also participated in the dialect identification shared tasks organized within the scope of the VarDial workshop series held from 2014 to 2017. The three related shared tasks organized at the VarDial workshop thus far are the Discriminating between Similar Languages (DSL) task organized from 2014 to 2017, Arabic Dialect Identification (ADI) organized in 2016 and 2017, and German Dialect Identification (GDI) organized in 2017.

Next we list some of the teams that adapted systems from NLI to dialect identification in the past few years.

\begin{itemize}
\item Variations of the string kernels method by the Unibuc team \cite{popescu-ionescu:2013:BEA8} competed in the ADI task in 2016 \cite{ionescu-popescu:2016:VarDial3} and in 2017 \cite{ionescu-butnaru:2017:VarDial} achieving the best results. 
\item Cologne-Nijmegen's TF-IDF-based approach \cite{gebre-EtAl:2013:BEA8} competed in the DSL shared task 2015 \cite{zampieri-EtAl:2015:LT4VarDial2} as team MMS ranking among the top 3 systems.
\item A variation of NRC's SVM approach \cite{goutte-leger-carpuat:2013:BEA8} competed in the DSL 2014 \cite{goutte-leger-carpuat:2014:VarDial} achieving the best results.
\item Bobicev applied Prediction for Partial Matching (PPM) in the NLI shared task \cite{bobicev:2013:BEA8} with results that did not reach top ten performance. A similar improved approached competed in the DSL 2015 \cite{bobicev:2015:LT4VarDial} ranking in the top half of the table.
\item A similar approach to the one by Jarvis \cite{jarvis-bestgen-pepper:2013:BEA8} that ranked 1\textsuperscript{st} place in the NLI task 2013 competed in the DSL 2017 \cite{bestgen:2017:VarDial}, achieving the best performance in the competition. 
\item Variations of MQ's SVM ensemble approach \cite{malmasi-wong-dras:2013:BEA8} have competed in the DSL 2015 \cite{malmasi2015dsl} and the ADI 2016 \cite{malmasi-zampieri:2016:VarDial3}, achieving the best performance in both shared tasks.

\end{itemize}

\noindent This section evidenced an important overlap between NLI methods and dialect identification methods both in terms of participation overlap in the shared tasks and in terms of successful approaches. With the exception of \newcite{bobicev:2013:BEA8}, most teams that were ranked among the top ten entries in the NLI shared task were also successful at the VarDial workshop shared tasks. 

Detailed information about all approaches and performance obtained in these competitions can be found in the VarDial shared task reports \cite{zampieri:2014:VarDial,zampieri:2015:LT4VarDial,dsl2016,vardial2017} and in the evaluation paper by \newcite{dslrec:2016}. 

\section{Methods}

In the next sections we describe the data provided by the shared task organizers and the ensemble SVM approach applied by the ZCD team.

\subsection{Data}

The organizers of the NLI Shared Task 2017 provided participants with data corresponding to eleven native languages: Arabic, Chinese, French, German, Hindi, Italian, Japanese, Korean, Spanish, Telugu and Turkish. The training dataset consists of 11,000 essays, orthographic transcriptions of 45-second English spoken responses, and iVectors (1,000 instances for each of the eleven native languages), while the development dataset was stratified similarly, containing 100 instances for each native language. 

There were individual tracks in which only the essays or only the responses could be used and a fusion track in which both the essays and the speech transcriptions (including iVectors) could be used. The test dataset, containing 1,100 instances with essays, speech transcriptions and iVectors, was released at a later date.

The use of a dataset containing text and speech is the main new aspect of the 2017 NLI task so we decide to compete in the fusion track taking both modalities into account. The approach used in our submission is described next.

\subsection{Approach}

\begin{table*}[!ht]
\center
\scalebox{0.98}{
\begin{tabular}{lll}
\hline
\bf System & \bf F1 (macro) & \bf Accuracy \\
\hline
Essays + Transcriptions + iVectors & 0.8358 & 0.8355 \\
Essays + Transcriptions & 0.8191 & 0.8191 \\
\hline
Official Baseline (with iVectors) & 0.7901 & 0.7909 \\
Official Baseline (without iVectors) & 0.7786 &  0.7791 \\
Random Baseline & 0.0909 & 0.0909 \\
\hline
\end{tabular}
}
\caption{ZCD results and baselines for the fusion track.}
\label{tab:results-FUSION-closed}
\end{table*}

We built a classification system based on SVM ensembles, following the methodology proposed by \newcite{malmasi2015dsl}.

The idea behind classification ensembles is to improve the overall performance by combining the results of multiple classifiers. Such systems have proved successful not only in NLI and dialect identification, as evidenced in the previous sections, but also in numerous text classification tasks, among which are complex word identification \cite{malmasi2016ltg} and grammatical error diagnosis \cite{xiang_et_al_2015}. The classifiers can differ in a wide range of aspects, such as algorithms, training data, features or parameters.

In our system, the classifiers used different features. We experimented with the following features: character $n$-grams (with $n$ in $\{1, ..., 10\}$) from essays and speech transcripts, word $n$-grams (with $n$ in $\{1, 2\}$) from essays and speech transcripts, and iVectors. For the n-gram features we used TF-IDF weighting applied on the tokenized version of the essays and speech transcripts (provided by the organizers). As a pre-processing step, we lowercased all words.

We first trained a classifier for each type of feature using the essays as input data, and performed cross-validation to determine the optimal value for the SVM hyperparameter $C$, searching in $\{10^{-5}, ..., 10^5\}$. Further, for the n-gram features we kept only those classifiers whose individual cross-validation performance was higher than 0.8. Thus, our first ensemble consisted of individual classifiers using character $n$-grams (with $n$ in $\{6, 7, 8\}$) from essays and speech transcripts. 

For the second ensemble, we introduced an additional classifier using the iVectors as features. To combine the classifiers, we employed a \mbox{majority-based} fusion method: the class label predicted by the ensemble is the one that was predicted by the majority of the classifiers. We used the SVM implementation provided by Scikit-learn \cite{scikit-learn}, based on the Liblinear library \cite{liblinear}.

On the development dataset, the first ensemble (essays + speech transcripts) obtained 0.83 accuracy, and the second ensemble (essays + speech transcripts + iVectors) obtained 0.84 accuracy.

\section{Results}
\label{sec:results}

We submitted two runs of our system. The first run included the essays and the transcriptions of responses, whereas the second run included also the iVectors. We present the results obtained by the two runs along with a random baseline and the performance of the unigram-based official baseline system in terms of F1 score and accuracy in Table \ref{tab:results-FUSION-closed}. 

The best results were achieved by the second run, reaching 83.55\% accuracy and 83.58\% F1 score. As can be seen in Table \ref{tab:results-FUSION-closed}, the iVectors bring a performance improvement of about 1.6 percentage points in terms of accuracy and F1 score.

Ten teams participated in the fusion track and our best run was ranked 3\textsuperscript{rd} by the shared task organizers. Ranks were calculated using McNemar’s test for statistical significance, a common practice in many NLI shared tasks (e.g. DSL 2016 \cite{dsl2016}, and the shared tasks at WMT \cite{bojar2016findings}).

The confusion matrix of our best submission is presented in Table \ref{tab:matrix}. We observed that the best performance was obtained for Japanese and the worst performance was obtained for Arabic. Not surprisingly, most confusion occurred between Hindi and Telugu. Our initial analysis indicates that this confusion occurred because of geographic proximity and not by intrinsic linguistic properties shared by these two languages, as Hindi and Telugu do not belong to the same language family - Hindi is a Hindustani language and Telugu is a Dravidian language.

\begin{table*}[!ht]
\center
\scalebox{0.96}{
\begin{tabular}{lrrrrrrrrrrr}
\hline
     &   CHI &   JPN &   KOR &   HIN &   TEL &   FRE &   ITA &   SPA &   GER &   ARA &   TUR \\
\hline
 CHI &    91 &     3 &     2 &     0 &     0 &     0 &     2 &     0 &     0 &     1 &     1 \\
 JPN &     2 &    93 &     2 &     0 &     1 &     1 &     0 &     0 &     1 &     0 &     0 \\
 KOR &     4 &    14 &    77 &     0 &     0 &     1 &     1 &     1 &     0 &     1 &     1 \\
 HIN &     1 &     0 &     1 &    80 &    18 &     0 &     0 &     0 &     0 &     0 &     0 \\
 TEL &     0 &     0 &     1 &    18 &    78 &     0 &     0 &     1 &     0 &     2 &     0 \\
 FRE &     2 &     0 &     0 &     2 &     1 &    87 &     5 &     0 &     2 &     1 &     0 \\
 ITA &     0 &     0 &     0 &     1 &     0 &     6 &    85 &     3 &     3 &     2 &     0 \\
 SPA &     1 &     1 &     2 &     2 &     1 &     4 &     7 &    77 &     2 &     2 &     1 \\
 GER &     0 &     1 &     0 &     3 &     0 &     3 &     2 &     1 &    90 &     0 &     0 \\
 ARA &     2 &     2 &     2 &     3 &     2 &     7 &     1 &     2 &     1 &    77 &     1 \\
 TUR &     1 &     2 &     0 &     3 &     0 &     2 &     3 &     1 &     1 &     3 &    84 \\
\hline
\end{tabular}
}
\caption{Confusion matrix on the test set.}
\vspace{4mm}
\label{tab:matrix}
\end{table*}





\section{Most Informative Features}

\begin{table*}[ht!]
\centering
\scalebox{0.93}{
    \begin{tabular}{ll}
    \hline
    \bf Language  & \bf Most Informative Features  \\ \hline
    
   Arabic & \textbar alot of \textbar\hphantom{ }alot of\textbar y thing \textbar\hphantom{ }statmen\textbar statment\textbar e alot o\textbar tatment \textbar ery thin\textbar very thi\textbar every th\textbar \\
    
    Chinese & \textbar\hphantom{ }i think\textbar\hphantom{ }taiwan \textbar i think \textbar beijing \textbar\hphantom{ }beijing\textbar\hphantom{ }taipei \textbar\hphantom{ }in chin\textbar in china\textbar n china \textbar\hphantom{ }chinese\textbar \\
    
    French & \textbar\hphantom{ }indeed \textbar . indeed\textbar\hphantom{ }. indee\textbar indeed ,\textbar ndeed , \textbar\hphantom{ }france \textbar developp\textbar\hphantom{ }french \textbar to concl\textbar o conclu\textbar \\
    
    German & \textbar\hphantom{ }, that \textbar\hphantom{ }and um \textbar\hphantom{ }germany\textbar germany \textbar\hphantom{ }berlin \textbar , that t\textbar have to \textbar\hphantom{ }have to\textbar\hphantom{ }um yeah\textbar um yeah \textbar \\
    
    Hindi & \textbar\hphantom{ }towards\textbar towards \textbar as compa\textbar\hphantom{ }as comp\textbar various \textbar\hphantom{ }various\textbar s compar\textbar\hphantom{ }enjoyme\textbar\hphantom{ }mumbai \textbar\hphantom{ }behind \textbar \\
    
    Italian & \textbar hink tha\textbar nk that \textbar ink that\textbar\hphantom{ }in ital\textbar n italy \textbar in fact \textbar in italy\textbar\hphantom{ }in fact\textbar i think \textbar\hphantom{ }italian\textbar \\
    
    Japanese & \textbar in japan\textbar n japan \textbar\hphantom{ }in japa\textbar apanese \textbar\hphantom{ }japanes\textbar japanese\textbar\hphantom{ }japan ,\textbar japan , \textbar i disagr\textbar\hphantom{ }japan .\textbar \\
    
    Korean & \textbar\hphantom{ }korean \textbar in korea\textbar\hphantom{ }in kore\textbar n korea \textbar\hphantom{ }however\textbar however \textbar korea , \textbar\hphantom{ }korea ,\textbar\hphantom{ }. howev\textbar . howeve\textbar \\
    
    Spanish & \textbar\hphantom{ }mexico \textbar oing to \textbar going to\textbar\hphantom{ }going t\textbar\hphantom{ }diferen\textbar le that \textbar\hphantom{ }the cit\textbar es that \textbar diferent\textbar ple that\textbar \\
    
    Telugu & \textbar ing the \textbar hyderaba\textbar\hphantom{ }hyderab\textbar yderabad\textbar derabad \textbar\hphantom{ }subject\textbar\hphantom{ }we can \textbar\hphantom{ }i concl\textbar i conclu\textbar where as\textbar \\
    
    Turkish & \textbar\hphantom{ }turkey \textbar istanbul\textbar stanbul \textbar\hphantom{ }istanbu\textbar\hphantom{ }uh and \textbar n turkey\textbar\hphantom{ }in turk\textbar in turke\textbar s about \textbar\hphantom{ }. becau\textbar \\

    \hline
    \end{tabular}
}
\caption{Top ten most informative character 8-grams for each language.}
\label{tab:features}
\end{table*}

As briefly discussed in the introduction of this paper, NLI methods can provide interesting information about patterns in non-native language that can be used to study second language acquisition and L1 interference or language transfer. For this purpose, in Table \ref{tab:features} we present the top ten most informative character 8-grams for each of the eleven languages in the dataset according to our classifier. 

It is not surprising that named entities are very informative for our system and highly discriminative for most native languages. For example, essays and responses from China often contain place names like {\em China, Taipei, Taiwan}, and {\em Beijing}, whereas those from Turkey contain {\em Istanbul} and, of course, {\em Turkey}. These features are very frequent in essays and responses by Chinese and Turkish speakers due to topical bias and not because of any intrinsic linguistic property of Chinese or Turkish. However, in other languages, interesting linguistic patterns can be identified by looking at these features. 

\noindent In the most informative features for French, for example, we find {\em developp} from the French {\em d\'{e}velopp\'{e}} which leads to a misspelling of the English word {\em developed}. In Arabic we observed a number of features that indicate misspellings. The Arabic alphabet is very different from the Latin one, making spelling English words particularly challenging for native speakers of Arabic. The top ten most informative features for Arabic include word boundary errors such as {\em every thing} for {\em everything}, and {\em alot} for {\em a lot}, as well as the omission of vowels such as {\em statment} for {\em statement}.

\section{Conclusion}

To the best of our knowledge, the NLI Shared Task 2017 fusion track was the first shared task to provide both written and spoken data for NLI. It was an interesting opportunity to evaluate the performance of NLI methods beyond written texts.

In this paper we highlighted the overlap between NLI and dialect, language variety, and similar language identification and used an approach that achieved high results in both tasks. We applied an SVM ensemble approach trained character $n$-grams achieving competitive results of 83.55\% accuracy ranking 3\textsuperscript{rd} in the fusion track.

Even though the results obtained by our approach were not low, we believe that there is still room for improvement. In previous shared tasks (e.g. NLI 2013, DSL 2015, and ADI 2016) we observed that SVM ensembles ranked higher in the results tables than our method did in the NLI 2017. We are investigating whether the combination of features or the implementation itself can be optimized for better performance.

\section*{Acknowledgements}
We would like to thank the NLI Shared Task 2017 organizers for making the dataset available and for replying promptly to all our inquiries. We further thank the anonymous reviewers for their valuable feedback.

Liviu P. Dinu is supported by UEFISCDI, project number 53BG/2016.

\bibliography{nli2017}
\bibliographystyle{emnlp_natbib}

\end{document}